%% file: main.tex
\documentclass[conference]{IEEEtran}
\IEEEoverridecommandlockouts
\usepackage{cite}
\usepackage{amsmath,amssymb,amsfonts}
\usepackage{algorithmic}
\usepackage{graphicx}
\usepackage{textcomp}
\usepackage{xcolor}
\usepackage{cite}
\usepackage{booktabs}
\usepackage{makecell}
\usepackage{multirow}
\usepackage{graphicx}
\usepackage{float}
\usepackage{url}
\usepackage{hyperref}
\def\BibTeX{{\rm B\kern-.05em{\sc i\kern-.025em b}\kern-.08em
    T\kern-.1667em\lower.7ex\hbox{E}\kern-.125emX}}
    
\makeatletter
\newcommand{\twocolumnfootnotefullwidth}[1]{%
  \begingroup
  \renewcommand{\thefootnote}{}
  \footnotetext{%
    \noindent\hspace*{-1em}\rule{0.3\linewidth}{0.4pt}\\[0.2em] 
    \vspace*{-1em}
    \noindent\footnotesize #1
  }
  \endgroup
}
\makeatother
\begin{document}

\title{MAESIL: Masked Autoencoder for Enhanced Self-supervised Medical Image Learning}

\author{
    \IEEEauthorblockN{Kyeonghun Kim}
    \IEEEauthorblockA{\textit{OUTTA} \\
                     kyeonghun.kim@outta.ai}
    \and
    \IEEEauthorblockN{Hyeonseok Jung}
    \IEEEauthorblockA{\textit{Chung-Ang University} \\
                     mjmk0820@cau.ac.kr}
    \and
    \IEEEauthorblockN{Youngung Han}
    \IEEEauthorblockA{\textit{Seoul National University} \\
                     yuhan@snu.ac.kr}
    \and
    \IEEEauthorblockN{Junsu Lim}
    \IEEEauthorblockA{\textit{Sangmyung University} \\
                     202115042@sangmyung.ac.kr}
    \and
    
    \IEEEauthorblockN{YeonJu Jean}
    \IEEEauthorblockA{\textit{Ewha Womans University} \\
                     ahxlzjt@ewhain.net}
    \and
    \IEEEauthorblockN{Seongbin Park}
    \IEEEauthorblockA{\textit{Seoul National University} \\
                     tjdqls@snu.ac.kr}
    \and
    \IEEEauthorblockN{Eunseob Choi}
    \IEEEauthorblockA{\textit{GIST} \\
                     eunseobchoi@gm.gist.ac.kr}
    \and
    \IEEEauthorblockN{Hyunsu Go}
    \IEEEauthorblockA{\textit{Seoul National University} \\
                     hsmail02@snu.ac.kr}
    \and

    \IEEEauthorblockN{Seoyoung Ju}
    \IEEEauthorblockA{\textit{Sangmyung University} \\
                     202115055@sangmyung.ac.kr}
    \and
    \IEEEauthorblockN{Seohyoung Park}
    \IEEEauthorblockA{\textit{Ewha Womans University} \\
                     03nobel@ewhain.net}
    \and
    \IEEEauthorblockN{Gyeongmin Kim}
    \IEEEauthorblockA{\textit{Chung-Ang University} \\
                     rnrn6442@cau.ac.kr}
    \and
    \IEEEauthorblockN{MinJu Kwon}
    \IEEEauthorblockA{\textit{Chung-Ang University} \\
                     kmjcap4@cau.ac.kr}
    \and

    \IEEEauthorblockN{Kyungseok Yuh}
    \IEEEauthorblockA{\textit{Dankook University} \\
                     32192743@dankook.ac.kr}
    \and
    \IEEEauthorblockN{Soo Yong Kim}
    \IEEEauthorblockA{\textit{AI Matics} \\
                     ksyint@aimatics.ai}
    \and
    \IEEEauthorblockN{Ken Ying-Kai Liao}
    \IEEEauthorblockA{\textit{NVIDIA ATC, Taiwan} \\
                     kenyingkail@nvidia.com}
    \and
    
    
    \IEEEauthorblockN{Nam-Joon Kim\textsuperscript{\dag}}
    \IEEEauthorblockA{\textit{Seoul National University} \\
                     knj01@snu.ac.kr}
    \and
    \IEEEauthorblockN{Hyuk-Jae Lee}
    \IEEEauthorblockA{\textit{Seoul National University} \\
                     hjlee@capp.snu.ac.kr}
}

\maketitle
\twocolumnfootnotefullwidth{\quad {\dag} Corresponding author}

    \input{00_abstract.tex}
    \input{01_introduction.tex}

    \input{02_related_work.tex}
    \input{03_methods}
    \input{04_experiments}
    \input{05_conclusion}
    \input{06_ak.tex}

\vspace{12pt}

\bibliographystyle{IEEEtran}
\bibliography{references}

\end{document}

%% file: 00_abstract.tex
\begin{abstract}
Training deep learning models for three-dimensional (3D) medical imaging, such as Computed Tomography (CT), is fundamentally challenged by the scarcity of labeled data. While pre-training on natural images is common, it results in a significant domain shift, limiting performance. Self-Supervised Learning (SSL) on unlabeled medical data has emerged as a powerful solution, but prominent frameworks often fail to exploit the inherent 3D nature of CT scans. These methods typically process 3D scans as a collection of independent 2D slices, an approach that fundamentally discards critical axial coherence and the 3D structural context. To address this limitation, we propose the autoencoder for enhanced self-supervised medical image learning(MAESIL), a novel self-supervised learning framework designed to capture 3D structural information efficiently. The core innovation is the `superpatch,' a 3D chunk-based input unit that balances 3D context preservation with computational efficiency. Our framework partitions the volume into superpatches and employs a 3D masked autoencoder strategy with a dual-masking strategy to learn comprehensive spatial representations. We validated our approach on three diverse large-scale public CT datasets. Our experimental results show that MAESIL demonstrates significant improvements over existing methods such as AE, VAE and VQ-VAE in key reconstruction metrics such as PSNR and SSIM. This establishes MAESIL as a robust and practical pre-training solution for 3D medical imaging tasks.

\end{abstract}

\begin{IEEEkeywords}

Self-Supervised Learning, Medical Imaging, Masked Autoencoder, 3D CT, Representation Learning
\end{IEEEkeywords}

\begin{figure*}[t!]
    \centering
    \includegraphics[width=0.9\textwidth]{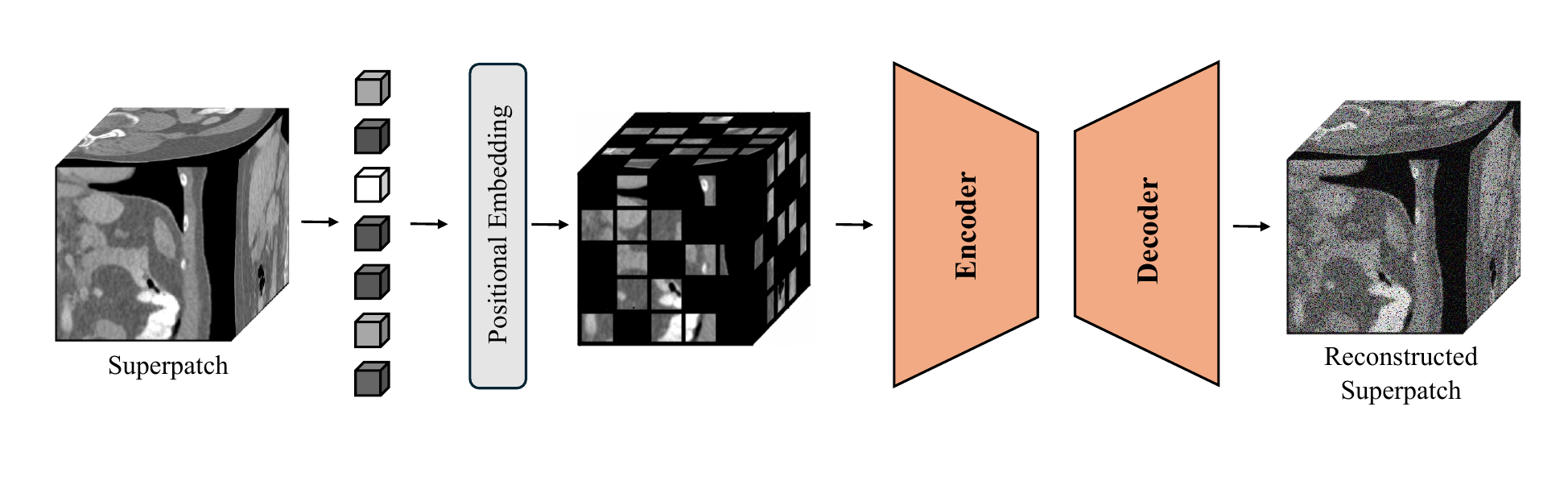}
    \caption{Overview of the proposed MAESIL framework, which masks a high ratio of an input 3D superpatch, feeds only the visible portions to an encoder, and uses a decoder to reconstruct the original superpatch from the encoded information and learnable mask tokens.}
    \label{fig:fig1_label}
\end{figure*}

%% file: 01_introduction.tex
\section{Introduction}

Deep learning techniques in medical imaging are rapidly becoming essential for clinical diagnostics and treatment planning~\cite{litjens2017survey}. Among various imaging modalities, Computed Tomography (CT) is widely utilized as it provides three-dimensional (3D) scans of complex anatomical structures within the human body, making it indispensable for comprehensive disease detection.

However, fully leveraging the potential of these CT scans with deep learning faces several fundamental challenges.
First, the \textbf{scarcity of labeled data} remains the most significant barrier to training deep learning models~\cite{huang2023self}. The expert annotation process required for creating large-scale datasets is exceptionally time-consuming and costly.
Second, to circumvent this issue, models pre-trained on natural images (e.g., ImageNet \cite{imagenet}) are often used. However, this approach leads to a \textbf{domain shift} problem, as the visual representations learned from natural images are fundamentally different from the unique characteristics of CT imaging, resulting in limited performance~\cite{guan2022domain}.

To address these core problems, Self-Supervised Learning (SSL) has emerged as a powerful alternative that utilizes large-scale unlabeled data~\cite{gao2023comparing}. In particular, pre-training on domain-specific datasets (i.e., medical images) rather than natural images has proven effective in mitigating the domain shift problem, demonstrating the strong potential of this direction.

However, these preceding approaches do not fully exploit the potential of 3D CT volumes, which is the inherent nature of CT scans. Many existing methods, including prominent SSL frameworks, treat the 3D CT scans as a mere collection of independent 2D slices for pre-training. While this 2D-based approach may be computationally convenient, it has an intrinsic limitation: it fails to capture the axial coherence and 3D structural context between slices, which are critical for accurate CT analysis~\cite{singh2023leveraging}.

The goal of this paper is to overcome this critical limitation of 2D-based SSL approaches.
We propose a novel self-supervised learning framework that introduces a new 3D input unit capable of capturing 3D structural information efficiently.
This approach avoids the information loss seen in 2D slice methods while mitigating the prohibitive computational burden associated with processing entire 3D scans, enabling a practical and effective pre-training solution for 3D CT scans.

%% file: 02_related_work.tex
\section{Related Work}

Masked Autoencoders (MAE) have recently emerged as one of the most scalable and effective self-supervised learning (SSL) frameworks for visual representation learning~\cite{he2021masked}. Inspired by masked language modeling, MAE learns rich representations by reconstructing randomly masked patches from a small subset of visible patches~\cite{devlin2018bert}. This powerful paradigm has naturally been extended to the medical imaging domain to address its unique challenges.

A prominent example is \textbf{MedMAE}, which directly tackles the critical domain shift problem~\cite{gupta2024medmae}. The authors of MedMAE correctly identified that models pre-trained on natural images (e.g., ImageNet) perform poorly on medical tasks.To solve this, they demonstrated that pre-training on a large-scale, diverse, unlabeled dataset (MID) allows the model to learn domain-specific visual representations.
Their results convincingly show that this domain-specific backbone significantly outperforms ImageNet-trained models across various downstream tasks, including classification and segmentation.

However, while MedMAE successfully addresses the domain shift problem, its methodology for handling 3D data like CT reveals a critical limitation.
Although their extensive MID dataset includes 3D modalities such as CT and MRI, the framework itself processes these complex volumes as a collection of independent 2D slices.
This 2D-based approach, while computationally convenient, fundamentally discards the 3D structural context and axial coherence between slices. This inter-slice information is essential for a comprehensive understanding of anatomical structures and pathologies in CT analysis.

As the work on MedMAE illustrates, the value of domain-specific pre-training is well-established. Yet, a methodology that effectively and efficiently captures the intrinsic 3D contextual information of CT scans, without resorting to 2D-slice simplification, remains an open challenge in the field.

%% file: 03_methods.tex
\section{Methods}
The overall architecture of our proposed MAESIL framework is
illustrated in Fig.~\ref{fig:fig1_label}. The central innovation of MAESIL is its input processing pipeline, which introduces a 3D chunk-based unit termed the `superpatch'. This design was conceived to balance two competing objectives: (1) mitigating the `3D context loss' from 2D slice-based methods and (2) managing the `high computational cost' of processing full 3D volumes.

\subsection{Superpatching and Masking}

\begin{figure}[htbp]
    \centering
    \includegraphics[width=0.9\linewidth]{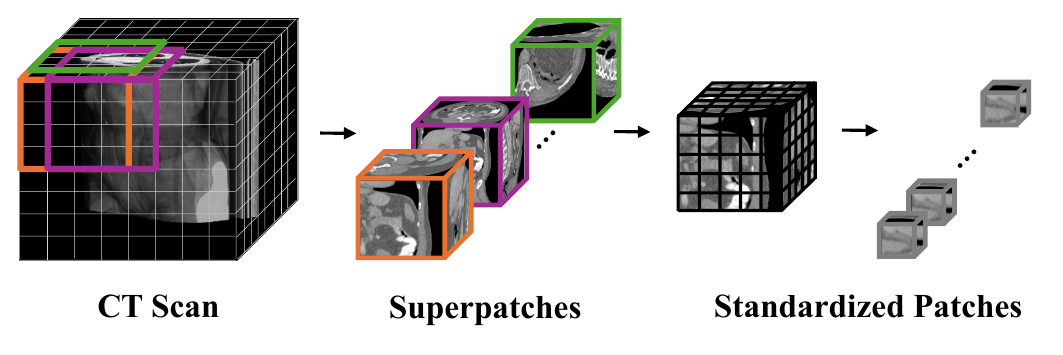}
    \caption{The 3D input processing pipeline. 
         A full 3D CT scans is first partitioned into `superpatches'. Each superpatch is then densely tokenized into smaller `standardized patches', which serve as the input tokens for the Transformer.}
    \label{fig:fig2_label}
\end{figure}

As illustrated in Fig.~\ref{fig:fig2_label}, an input CT scans, assumed to be $512\times512\times512$, is first processed by our model.
To balance 3D context against computational load, we partition this volume into 64 non-overlapping superpatches, each with dimensions of $128 \times 128 \times 128$.
This superpatch unit is then further tokenized.
It is densely divided into $8 \times 8 \times 8$ local patches, yielding 4,096 tokens per superpatch.
A 3D convolution layer embeds these local patches into a latent vector space (e.g., 768-dimensional) suitable for the Transformer.
To preserve spatial relationships, positional encodings are added to each token.

For the pre-training objective, we adapt the MAE strategy for 3D data.
A dual-masking strategy is applied to the superpatch tokens.
First, `plane-wise' masking randomly discards 75\% of the patches within each 2D plane.
Second, `axis-wise' masking removes a contiguous block of 50\% of the slices along the Superior-Inferior (S-I) axis.

\subsection{Encoder-Decoder Architecture}
The encoder-decoder architecture is responsible for reconstructing the original superpatch from the masked input.

The decoder first receives the compressed embeddings from the encoder. Concurrently, learnable mask tokens are inserted into the original positions of the masked patches to complete the full sequence. These mask tokens serve as signals, informing the decoder which positions need to be reconstructed.

This full sequence is then reordered to its original order and passes through a stack of Transformer blocks within the decoder. The Transformer blocks learn the contextual relationships within the sequence via self-attention. Through this process, predictions for each patch are made. Finally, all predicted patches are reassembled into a single superpatch, restoring the original volume.

%% file: 04_experiments.tex
\section{Experiments and Results}

\subsection{Dataset}
Our model's performance and generalization capabilities are validated across three distinct, large-scale, publicly available medical CT datasets: BTCV, LIDC-IDRI, and TotalSegmentator. These datasets were chosen to cover a wide spectrum of anatomical regions, from abdominal organs and thoracic scans to a full-body. A summary of these datasets is provided in Table~\ref{tab:datasets}.

\begin{table}[H]
    \centering
    \caption{Overview of the datasets used for pre-training.}
    \label{tab:datasets}
    \begin{tabular}{lccc}
        \toprule
        Dataset & \# Scans & Region \\
        \midrule
        BTCV~\cite{BTCV} & 30 & Abdominal \\
        LIDC-IDRI~\cite{LIDC-IDRI} & 1,018 & Thoracic \\
        TotalSegmentator~\cite{wasserthal2023totalsegmentator} & 1,204 & Full-body\\
        \bottomrule
    \end{tabular}
\end{table}

\textbf{BTCV}~\cite{BTCV} The ``Multi-Atlas Labeling Beyond the Cranial Vault" (BTCV) dataset is a standard benchmark for abdominal organ segmentation. It contains 30 subjects with contrast-enhanced abdominal CT scans and provides high-quality man-ual segmentations for 13 abdominal structures.

\textbf{LIDC-IDRI}~\cite{LIDC-IDRI} The ``Lung Image Database Consortium and Image Database Resource Initiative" (LIDC-IDRI) is a large-scale public database of thoracic CT scans. It consists of 1,018 cases, with detailed XML annotations for lung nodules (lesions) independently delineated by four experienced radiologists.

\textbf{TotalSegmentatorV2}~\cite{wasserthal2023totalsegmentator} This is a recent, large-scale benchmark for robust, segmentation, containing 1,204 CT examinations~\cite{guo2025maisi}. Its key feature is the comprehensive annotation of 104 different anatomical structures (including 27 organs, 59bones, 10 muscles, and 8 vessels). The dataset is intentionally diverse, representing a wide variety of patient ages, scanners, and pathologies. While this diversity poses a significant challenge, it is precisely what makes the dataset ideal for evaluating model robustness to varied scan protocols and patient conditions.

For our pre-training, these three distinct collections were combined into a single, unified dataset to force the model to learn generalized representations across all domains.

\subsection{Comparison Results}
To evaluate the reconstruction performance of our proposed MAESIL framework, we conduct a comprehensive quantitative comparison against several standard generative baseline models. These baselines include a standard Autoencoder (AE), a Variational Autoencoder (VAE), and a Vector Quantized Variational Autoencoder (VQ-VAE)~\cite{hinton2006reducing, kingma2013auto, van2017neural}. While our MAESIL is trained on a masked reconstruction (inpainting) task, the baselines are trained on the standard full reconstruction task, i.e., reconstructing the original uncorrupted input. To rigorously test generalization, all models were pre-trained under fair conditions on a single, unified dataset that combines all three previously mentioned collections (BTCV, LIDC-IDRI, and TotalSegmentator). This forced the models to learn robust representations from a highly diverse mix of anatomical structures, body parts, and scan protocols. The performance is measured using three widely accepted metrics for image reconstruction quality: Peak Signal-to-Noise Ratio (PSNR), Structural Similarity Index Measure (SSIM), and Learned Perceptual Image Patch Similarity (LPIPS). For PSNR and SSIM, higher values indicate better reconstruction fidelity, while for the perceptually-based LPIPS metric, lower values signify that the reconstructed image is closer to the original in human perception. The consolidated results of this comparison are presented in Table~\ref{tab:comparison}.

\begin{table} [H]
    \centering
    \caption{Comparison with other models.}
    \label{tab:comparison}
    \begin{tabular}{lccc}
        \toprule
        Model & PSNR ($\uparrow$) & SSIM ($\uparrow$) & LPIPS ($\downarrow$) \\
        \midrule
        AE~\cite{hinton2006reducing} & 25.23 & 0.97 & 0.39 \\
        VAE~\cite{kingma2013auto} & 11.91 & 0.25 & 0.91 \\
        VQ-VAE~\cite{van2017neural} & 20.04 & 0.88 & 0.49 \\
        \midrule
        MAESIL (Ours) & \textbf{30.28} & \textbf{0.98} & \textbf{0.26} \\
        \bottomrule
    \end{tabular}
\end{table}

The results clearly demonstrate the superiority of our MAESIL
framework in all evaluation metrics. MAESIL achieves the
highest PSNR score of \textbf{30.28}, significantly surpassing
all baselines and indicating a much higher pixel-level accuracy
in our reconstructions. Similarly, in terms of structural
fidelity, our model obtains a near-perfect SSIM score of
\textbf{0.99}. The VAE, in particular, shows a notable
failure in capturing structural information, scoring only 0.25
on this metric. Most critically, MAESIL achieves the
\textbf{lowest LPIPS score of 0.26}, compared to 0.39 (AE),
0.49 (VQ-VAE), and 0.91 (VAE). This low perceptual distance
score confirms that our model's reconstructions are not only
pixel-accurate but also perceptually more realistic and less
blurrier than those from the baseline methods. This strong
performance validates the effectiveness of our 3D superpatch
and a dual-masking strategy in learning comprehensive 3D
representations.

\subsection{Discussion}
In this section, we analyze the quantitative and qualitative results of our framework. 

\begin{table}[H]
    \centering
    \caption{PSNR performance of our model (MAESIL) across different datasets.}
    \label{tab:dataset_ablation}
    \begin{tabular}{lc}
        \toprule
        Dataset & PSNR ($\uparrow$) \\
        \midrule
        BTCV ~\cite{BTCV}& 32.09 \\
        LIDC-IDRI ~\cite{LIDC-IDRI}& 22.58 \\
        TotalSegmentatorV2 ~\cite{wasserthal2023totalsegmentator} & 18.72 \\
        \bottomrule
    \end{tabular}
\end{table}
Table~\ref{tab:dataset_ablation} presents the quantitative results, showing the PSNR performance of our model when evaluated on each of the three datasets individually. The performance varies, achieving the highest PSNR of 32.09 on BTCV and the lowest of 18.72 on TotalSegmentator. This variation is expected; the TotalSegmentatorV2 dataset is by far the most diverse, mixing 104 anatomical structures, various body parts, and different scan protocols, which makes achieving a high average PSNR highly challenging. Critically, these results were obtained using a total masking ratio of 75\% and an embedding dimension of 768. This high ratio was implemented via our dual-masking scheme, which includes a difficult 50\% axial-wise masking. This aggressive masking strategy makes the inpainting task extremely challenging and prevents the model from trivially copying visible patches. The fact that MAESIL still achieves robust PSNR scores under this difficult condition strongly suggests that the model is forced to, and successfully does, learn meaningful 3D contextual representations from the limited visible data.

\begin{figure} [H]
    \centering
    \includegraphics[width=1.0\linewidth]{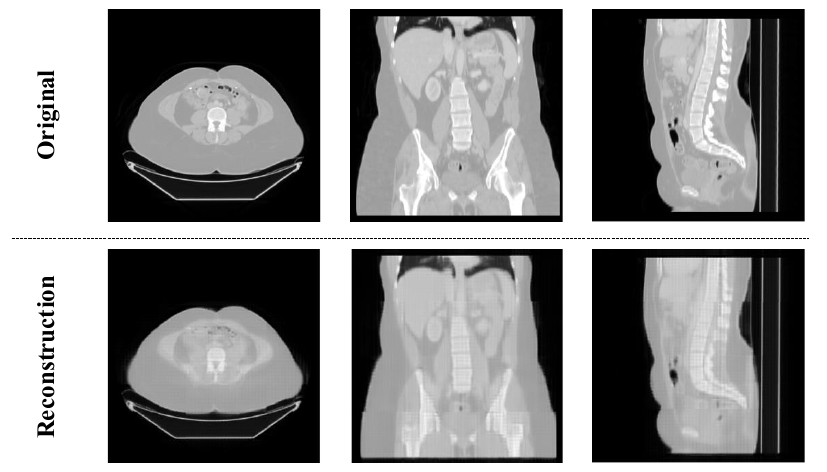}
    \caption{Qualitative reconstruction results of MAESIL across different anatomical views. The original input and reconstruction output are compared. From left to right: Axial, Coronal and Sagittal views.}
    \label{fig:fig3_label}
\end{figure}
Figure \ref{fig:fig3_label} provides a qualitative comparison of our reconstruction results across three different anatomical views (Axial, Coronal, and Sagittal). As shown, MAESIL successfully reconstructs complex anatomical structures—such as the vertebrae and various soft tissues—with high fidelity to the original input across all three axes. This visually confirms that our model effectively learns and preserves 3D contextual information rather than treating slices independently. However, fixed-pattern artifacts are observable in the reconstruction outputs. This is a common phenomenon in patch-based reconstruction methods related to the decoder's upsampling process, and addressing these artifacts is a clear direction for future work~\cite{he2021masked}.

%% file: 05_conclusion.tex
\section{Conclusion}
In this paper, we proposed MAESIL, a novel self-supervised learning framework based on a superpatch input unit and a dual-masking strategy to address the 3D structural context loss in existing 2D-based methods. Experiments on a unified dataset of three public benchmarks showed that MAESIL significantly outperforms standard reconstruction baselines like AE and VQ-VAE across all key metrics. This validates MAESIL as an effective pre-training strategy for 3D medical data. Future work will focus on refining the decoder architecture to reduce reconstruction artifacts and applying the pre-trained backbone to downstream tasks such as segmentation and classification.

%% file: 06_ak.tex
\section{Acknowledgment}
 This work was supported by the Next Generation Semiconductor Convergence and Open Sharing System, and by the Institute of Information \& Communications Technology Planning \& Evaluation (IITP) through the Artificial Intelligence Semiconductor Support Program to Nurture the Best Talents (IITP-2023-RS-2023-00256081), funded by the Ministry of Science and ICT of Korea (MSIT).

%% file: references.bib
@article{litjens2017survey,
  title={A survey on deep learning in medical image analysis},
  author={Litjens, Geert and Kooi, Thijs and Bejnordi, Babak Ehteshami and Setio, Arnaud Arindra Adiyoso and Ciompi, Francesco and Ghafoorian, Mohsen and Van Der Laak, Jeroen AWM and Van Ginneken, Bram and S{\'a}nchez, Clara I},
  journal={Medical Image Analysis},
  volume={42},
  pages={60--88},
  year={2017},
  publisher={Elsevier}
}

@article{huang2023self,
  title={Self-supervised learning for medical image classification: a systematic review and implementation guidelines},
  author={Huang, Shih-Cheng and Pareek, Anuj and Jensen, Malte and Lungren, Matthew P and Yeung, Serena and Chaudhari, Akshay S},
  journal={npj Digital Medicine},
  volume={6},
  number={1},
  pages={74},
  year={2023},
  publisher={Nature Publishing Group UK London}
}

@article{guan2022domain,
  title={Domain adaptation for medical image analysis: A survey},
  author={Guan, Hu and Liu, Ming-Hui},
  journal={IEEE Transactions on Biomedical Engineering},
  volume={69},
  number={7},
  pages={2245--2259},
  year={2022},
  publisher={IEEE}
}

@article{gao2023comparing,
  title={Comparing 3D, 2.5 D, and 2D Approaches to Brain Image Auto-Segmentation},
  author={Gao, Elliot L and Mark, William K and McKinney, A Isaiah and Xiao, Jian and Goldman, David A and Holcomb, Jonathan M and Young, Robert J},
  journal={Tomography},
  volume={9},
  number={2},
  pages={478--489},
  year={2023},
  publisher={MDPI}
}

@article{singh2023leveraging,
  title={Leveraging 2D deep learning ImageNet-trained models for native 3D medical image analysis},
  author={Singh, Swati and Kencha, Prathik and Prakash, GYAN},
  journal={Scientific Reports},
  volume={13},
  number={1},
  pages={18803},
  year={2023},
  publisher={Nature Publishing Group UK London}
}

@article{he2021masked,
  title={Masked autoencoders are scalable vision learners},
  author={He, Kaiming and Chen, Xinlei and Xie, Saining and Li, Yanghao and Doll{\'a}r, Piotr and Girshick, Ross},
  journal={arXiv preprint arXiv:2111.06377},
  year={2021}
}

@article{devlin2018bert,
  title={Bert: Pre-training of deep bidirectional transformers for language understanding},
  author={Devlin, Jacob and Chang, Ming-Wei and Lee, Kenton and Toutanova, Kristina},
  journal={arXiv preprint arXiv:1810.04805},
  year={2018}
}

@article{gupta2024medmae,
  title={MedMAE: A Self-Supervised Backbone for Medical Imaging Tasks},
  author={Gupta, Anubhav and Osman, Islam and Shehata, Mohamed S and Braun, John W},
  journal={arXiv preprint arXiv:2407.14784},
  year={2024}
}

@article{BTCV,
  author    = {Landman, Bennett A. and Xu, Zhoubing and others},
  title     = {BTCV: Multi-Atlas Labeling Beyond the Cranial Vault (Synapse)},
  journal   = {Synapse Repository},
  year      = {2015}
}

@article{LIDC-IDRI,
  author    = {Armato III, Samuel G. and others},
  title     = {LIDC/IDRI Collection (TCIA)},
  journal   = {The Cancer Imaging Archive},
  year      = {2015}
}

@article{wasserthal2023totalsegmentator,
  title={TotalSegmentator: robust segmentation of 104 anatomic structures in CT images},
  author={Wasserthal, Jakob and Breit, Hanns-Christian and Meyer, Manfred T and Pradella, Maurice and Hinck, Daniel and Sauter, Alexander W and Heye, Tobias and Boll, Daniel T and Cyriac, Joshy and Yang, Shan and others},
  journal={Radiology: Artificial Intelligence},
  volume={5},
  number={5},
  pages={e230024},
  year={2023},
  publisher={Radiological Society of North America}
}

@inproceedings{guo2025maisi,
  title={Maisi: Medical ai for synthetic imaging},
  author={Guo, Pengfei and Zhao, Can and Yang, Dong and Xu, Ziyue and Nath, Vishwesh and Tang, Yucheng and Simon, Benjamin and Belue, Mason and Harmon, Stephanie and Turkbey, Baris and others},
  booktitle={2025 IEEE/CVF Winter Conference on Applications of Computer Vision (WACV)},
  pages={4430--4441},
  year={2025},
  organization={IEEE}
}

@article{hinton2006reducing,
  title={Reducing the dimensionality of data with neural networks},
  author={Hinton, Geoffrey E and Salakhutdinov, Ruslan R},
  journal={science},
  volume={313},
  number={5786},
  pages={504--507},
  year={2006},
  publisher={American Association for the Advancement of Science}
}

@article{kingma2013auto,
  title={Auto-encoding variational bayes},
  author={Kingma, Diederik P and Welling, Max},
  journal={arXiv preprint arXiv:1312.6114},
  year={2013}
}

@article{van2017neural,
  title={Neural discrete representation learning},
  author={Van Den Oord, Aaron and Vinyals, Oriol and others},
  journal={Advances in neural information processing systems},
  volume={30},
  year={2017}
}

@article{imagenet,
  title={Imagenet large scale visual recognition challenge},
  author={Russakovsky, Olga and Deng, Jia and Su, Hao and Krause, Jonathan and Satheesh, Sanjeev and Ma, Sean and Huang, Zhiheng and Karpathy, Andrej and Khosla, Aditya and Bernstein, Michael and others},
  journal={International journal of computer vision},
  volume={115},
  number={3},
  pages={211--252},
  year={2015},
  publisher={Springer}
}
